\documentclass[10pt,twocolumn,letterpaper]{article}

\usepackage{authblk}

\usepackage[pagenumbers]{cvpr} 

\usepackage[pdftex]{graphicx}
\usepackage{grffile} 
\usepackage{amsmath}
\usepackage{amssymb}
\usepackage{booktabs}
\usepackage{bbm}
\usepackage{xspace}
\usepackage[export]{adjustbox}
\usepackage[pagebackref,breaklinks,colorlinks]{hyperref}

\DeclareMathOperator*{\argmin}{argmin}

\usepackage[capitalize]{cleveref}
\crefname{section}{Sec.}{Secs.}
\Crefname{section}{Section}{Sections}
\Crefname{table}{Table}{Tables}
\crefname{table}{Tab.}{Tabs.}

\def\cvprPaperID{8193} 
\def\confName{CVPR}
\def\confYear{2023}

\begin{document}

\title{Latent SHAP: Toward Practical Human-Interpretable Explanations}



\author[1]{Ron Bitton$^{*}$}
\author[1]{Alon Malach$^{*}$}
\author[1]{Amiel Meiseles$^{*}$}
\author[2]{Satoru Momiyama}
\author[2]{Toshinori Araki}
\author[2]{Jun Furukawa}
\author[1]{Yuval Elovici}
\author[1]{Asaf Shabtai}

\affil[1]{Ben-Gurion University of the Negev, Beer Sheva, Israel}
\affil[2]{NEC Corporation, Japan}
\maketitle
\def\thefootnote{*}\footnotetext{These authors contributed equally to this work}\def\thefootnote{\arabic{footnote}}
\begin{abstract}
Model agnostic feature attribution algorithms (\eg, SHAP and LIME) are ubiquitous techniques for explaining the decisions of complex classification models, such as deep neural networks.  
However, since complex classification models produce superior performance when trained on low-level (or encoded) features, in many cases, the explanations generated by these algorithms are neither interpretable nor usable by humans. 
Methods proposed in recent studies that support the generation of human-interpretable explanations are impractical, because they require a fully invertible transformation function that maps the model's input features to the human-interpretable features. 
In this work, we introduce Latent SHAP, a black-box feature attribution framework that provides human-interpretable explanations, without the requirement for a fully invertible transformation function.
We demonstrate Latent SHAP's effectiveness using (1) a controlled experiment where invertible transformation functions are available, which enables robust quantitative evaluation of our method, and (2) celebrity attractiveness classification (using the CelebA dataset) where invertible transformation functions are not available, which enables thorough qualitative evaluation of our method.
\end{abstract}

\section{\label{sec:intro}Introduction}
\begin{figure}
     \centering
     \begin{subfigure}[t]{0.361\textwidth}
         \centering
         \includegraphics[width=\textwidth]{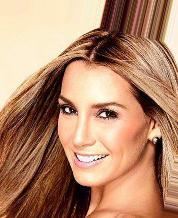}
         \caption{Original image}
         \label{fig:intro_a}
     \end{subfigure}
     \hfill
     \begin{subfigure}[t]{0.372\textwidth}
         \centering
         \includegraphics[width=\textwidth]{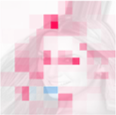}
         \caption{Naive feature importance generated using SHAP}
         \label{fig:intro_b}
     \end{subfigure}
     \hfill
     \begin{subfigure}[t]{0.48\textwidth}
         \centering
         \includegraphics[width=\textwidth]{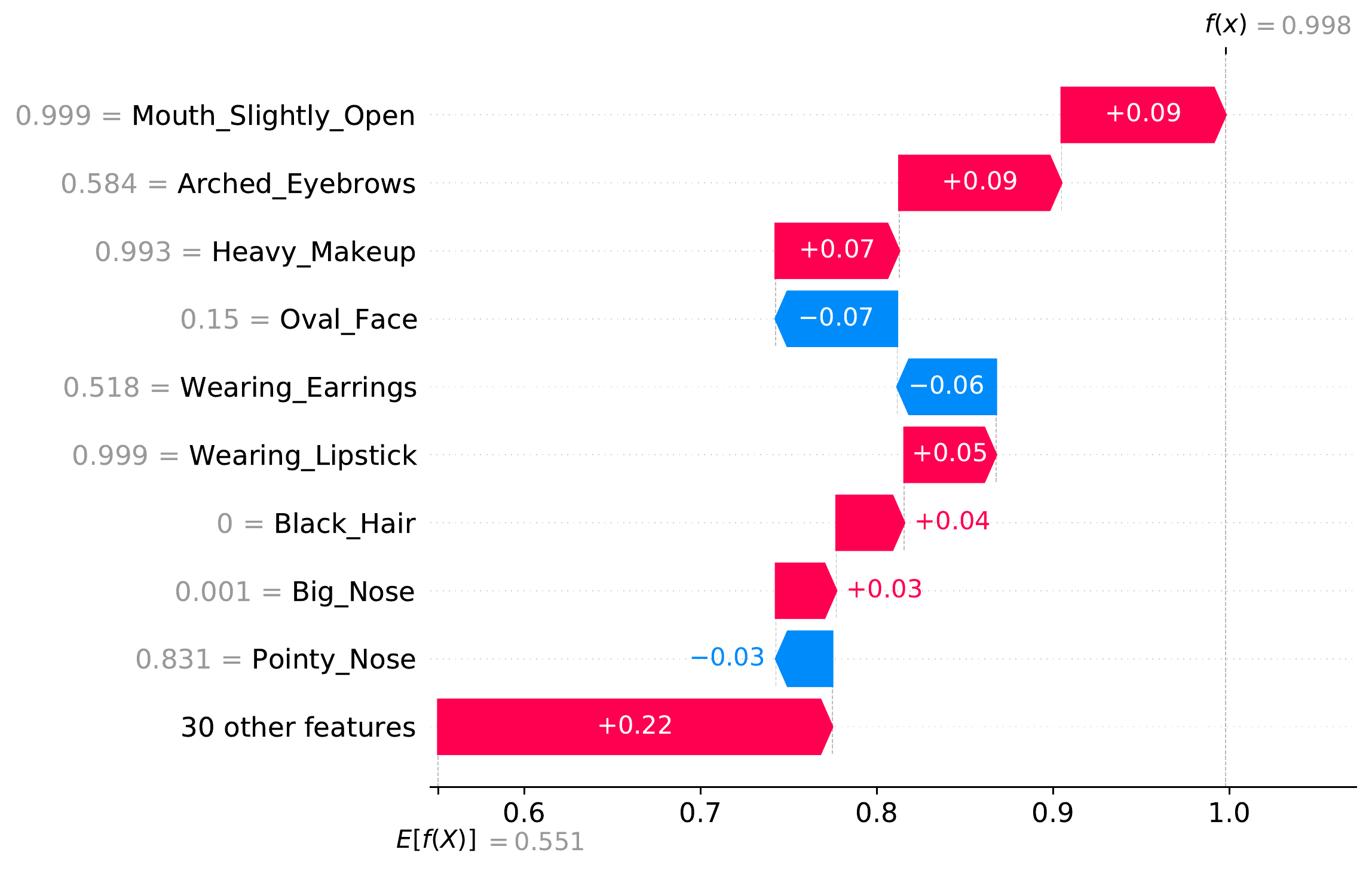}
         \caption{Human-interpretable explanations generated using Latent SHAP}
         \label{fig:intro_c}
     \end{subfigure}
        \caption{Explaining the decision of attractiveness classifier}
        \label{fig:intro}
\end{figure}

Deep neural networks (DNNs) have demonstrated superior performance in solving complex problems in numerous domains, including image classification~\cite{krizhevsky2012imagenet,he2015delving},
intrusion detection~\cite{buczak2015survey,vinayakumar2019robust}, drug discovery~\cite{segler2018generating}, natural language processing~\cite{brown2020language,devlin2018bert}
and fraud detection~\cite{roy2018deep}.
When humans use complex machine learning (ML) models to make crucial decisions, a vital concern is whether humans can trust the model and its predictions. 
This requirement is mandatory in many domains such as healthcare (for medical treatment decisions~\cite{amann2020explainability}), finance (for automated lending decisions~\cite{sachan2020explainable}), and autonomous vehicles (for decision-making by self-driving cars~\cite{Kim_2017_ICCV}) where human lives depend on the decisions made by the ML model.
Unfortunately, complex models such as DNNs are difficult to explain~\cite{rudin2019stop}.

Recent studies in the domain of explainable ML outline two desired characteristics for explainers: \textit{human interpretable} \ie, explanations must provide meaningful and qualitative understanding regarding the decision made by the model, by considering human's limitations, and \textit{faithfulness} \ie, explanations must correspond to how the model truly behaves.
Guided by these desired characteristics, recent research has proposed various methods for explaining ML models. These methods range from feature attribution approaches such as SHAP~\cite{lundberg2017unified} and LIME~\cite{ribeiro2016should}, concept-based attribution approaches such as TCAV~\cite{kim2018interpretability} and CaCE~\cite{goyal2019explaining}, example-based approaches such as counterfactual explanations~\cite{wachter2017counterfactual}, and rule-based approaches such as forest-based trees~\cite{sagi2020explainable} and DeepRED~\cite{zilke2016deepred}.

Among these approaches, \textit{local interpretable}, \textit{model-agnostic} feature attribution approaches (\eg, SHAP, LIME) are the most popular methods for explaining ML models.
These methods, train a second model that is inherently interpretable (such as a linear regression or decision tree model) on samples drawn from the local environment of the instance being predicted by the original model. That is, the second model is trained to approximate the decisions made by the original model \textit{in the local environment of the instance being explained}. The explanations resulting from the second model are at least \textit{locally faithful} to the decisions made by the original model~\cite{ribeiro2016should}.

The downside of feature attribution approaches is that in many use cases, they can be difficult to interpret by humans. 
Specifically, these approaches are designed to explain a model's decision with respect to the \emph{model's input features}. 
However, in many cases, the input features are raw or encoded, making it very difficult for humans (\eg, domain experts) to draw meaningful and useful conclusions. 

The abovementioned problem was not overlooked in recent works. 
Specifically, in the paper presenting LIME~\cite{ribeiro2016should}, the authors distinguish between model input features and interpretable data representations. 
Interpretable data representations must be understandable by humans, regardless of the actual features used by the model. 
In order to explain a model with respect to the interpretable data representations, the methods proposed in recent studies require the user to formulate an \textit{invertible} transformation function that maps the model's input features to the interpretable data representation~\cite{ribeiro2016should,de2020human}. 
We argue that a transformation of this type is not possible in many real-world use cases.

A motivating example in the domain of computer vision is explaining the predictions made by a facial attractiveness classifier~\cite{aarabi2001facial}. 
The input features of this type of classifier are raw three-colored pixels of an image of a person, and the output is whether or not this person is deemed attractive. 
In Figure~\ref{fig:intro}, we present pixel-level explanations generated using SHAP (b). 
As can be seen, such detailed explanations fail to provide meaningful insights regarding the attractiveness of the person in the original image (a).
In contrast, more meaningful insights can be obtained from an interpretable data representation, for instance, by using characteristics regarding a person's appearance, \eg, ``smiling," ``arched eyebrows," and ``wearing earrings" (c). 

Unfortunately, when using methods such as SHAP or LIME to derive explanations with respect to the interpretable data representation, one must formulate an invertible transformation function from the model's input features (\ie, raw three-colored pixels) to the interpretable data representation (\ie, characteristics regarding a person's appearance). 
In practice, formulating such a transformation function is very challenging, because there is no direct mapping from features in the abstract interpretable data representation to the very granular input features.

In this paper, we extend the framework presented in LIME and SHAP papers to support the generation of human-interpretable explanations, when an invertible transformation from the model's input features to the interpretable data representation is infeasible.
Latent SHAP, our proposed framework, can produce human-interpretable explanations when \emph{just a non-invertible transformation function from the model's input features to the interpretable data representation is available}. 

We evaluate Latent SHAP on (1) a toy example where invertible transformation functions are available, which enables us to perform a quantitative evaluation of the faithfulness of the proposed method with different hyperparameters and transformation functions; and (2) a real-world example from the computer vision domain - celebrity attractiveness classification (using the CelebA dataset) where invertible transformation functions are not available, which enables us to perform a qualitative evaluation of the faithfulness of the proposed framework in a real-world scenario.
\section{\label{sec:related}Feature Attribution Methods}
In this section, we review LIME and SHAP, which are the most popular locally interpretable model-agnostic feature attribution methods.

\subsection{LIME}
\underline{L}ocal \underline{I}nterpretable \underline{M}odel-agnostic \underline{E}xplanations (LIME) \cite{ribeiro2016should} explains the predictions $f(x)$ of a complex model $f$ for a given input sample $x$, using an explanation model $g\in G$, where $G$ is a class of interpretable models such as linear models or decision trees. 
In practice, LIME includes the following two main phases:

\noindent \textbf{\underline{Phase I:} Sampling a background dataset in the input's local environment.} 
The explanation model $g$ is trained to approximate $f$ in the local environment of $x$.
The locality of $x$ is represented using the proximity function $\pi_x$ as follows:
\begin{equation}
    \pi_x(z) = e^{-D(x,z)^2/\sigma^2}
    \label{eq:prox_function}
\end{equation}
\noindent where $D$ is a distance function (\eg, $L_2$ distance), and $\sigma$ is the width. 
The background dataset $B_x$ is sampled both in the vicinity of $x$ and far from $x$. 

\noindent\textbf{\underline{Phase II:} Training the explanation model.} The explanation model $g$ is trained to minimize the objective function:
\begin{equation}
\xi(x) = \argmin_{g \in G}  \mathcal{L}(g,f,\pi_x) + \Omega(g)
\label{eq:lime_optimization_function}
\end{equation}
\noindent where $\Omega(g)$ is a measure of the complexity of the explanation model (\eg, in linear models, $\Omega(g)$ may be the number of non-zero weights), and $\mathcal{L}(g,f,\pi_x)$ is a measure of how unfaithful $g$ is in approximating $f$ in the locality defined by $\pi_x$. 
$\mathcal{L}$ is defined as the locally weighted square loss:
\begin{equation}
\mathcal{L}(g,f,\pi_x)= \sum_{z\in B_x}  \pi_x(z)(f(z)-g(z))^2
\label{eq:lime_loss_function_for_noninterpretable_domain}
\end{equation}

\noindent\textbf{Using LIME to explain non-interpretable features.}  In the many cases where $x$ is not interpretable by humans, explainers such as LIME can also provide explanations with respect to other interpretable input $x'$ (a.k.a. simplified input). 
In such cases, the background dataset $B_{x'}$ is sampled from the local environment of $x'$ instead of $x$ using a proximity function $\pi_{x'}$. 
Furthermore, $g$ is trained to minimize the following loss function:
\begin{equation}
\mathcal{L}(g,f,\pi_{x'})=\sum_{z\in B_{x'}}  \pi_{x'}(z)(f(h^{-1}_x(z))-g(z))^2
\label{eq:lime_loss_function_for_interpretable_domain}
\end{equation}
where $x=h_x(x')$ is function that maps $x$ to $x'$.
Note that in order to produce explanations with respect to the interpretable domain, $g$ is trained on the local environment defined by $\pi_{x'}$, which is sampled from the local environment of the interpretable input $x'$.

\subsection{SHAP}\label{subsec:SHAP}
\underline{SH}apley \underline{A}dditive ex\underline{P}lanations (SHAP) \cite{lundberg2017unified} is a model-agnostic feature attribution method inspired by the concept of Shapley values from game theory. Shapley values provide a way of fairly dividing up a reward between the players of a game \cite{shapley1953}. The general idea of Shapley values is to estimate a player's contribution, using the reduction in the expected reward of all possible subsets of players that do not include them. When applied to explaining ML models, prediction is the game, the model output is the reward, and the features are the players.

SHAP (specifically Kernel SHAP) explains the predictions $f(x)$ of a complex model $f$ for a given input sample $x$, using an explanation model $g \in G$, where $G$ is a class of linear regression models. In practice, SHAP includes the following four main phases (see Figure \ref{fig:KernelShap}):

\noindent\textbf{\underline{Phase I:} Sampling a background dataset.} A background dataset $B$ is sampled from the distribution of the dataset used for training $f$.

\noindent\textbf{\underline{Phase II:} Creating coalition specific background datasets.} For each subset of features $c \subset \{x_1,x_2,...,x_n\}$ (denoted as a coalition of features), SHAP creates a coalition-specific background dataset $B^{c}_x$, in which values of features from the background dataset that are included in the coalition $c$ are replaced with their feature values from~$x$.

\noindent\textbf{\underline{Phase III:} Creating the dataset used to train the explanation model.} The dataset used for training $g$ (denoted as $B_x$) includes a single instance for each coalition-specific background dataset $B^{c}_x$. The features of this instance are binary indicators of the absence or presence of a feature within the coalition. 
The prediction value of this instance $\hat{y}^{c}_x$ is calculated by averaging the predictions of $f$ on $B^{c}_x$:
    \begin{equation}
        \hat{y}^{c}_x = \frac{1}{|B^{c}_x|}  * \sum_{z\in B^{c}_x}  f(z)
    \end{equation}
    
\noindent\textbf{\underline{Phase IV:} Training the explanation model.} Model $g$ is trained to minimize the following loss function:  
\begin{equation}
\mathcal{L}(g,f,\pi_x)= \sum_{c\in B_x}  \pi_x(c)(\hat{y}^c_x-g(c))^2
\label{eq:shap_loss_function_for_noninterpretable_domain}
\end{equation}
where $\pi_x(c)$ is a function defining the weight of a coalition $c$ as a function of its size $|c|$ as follows:
\begin{equation}
    \pi_x(c) = \frac{|c|!(n-|c|-1)!}{n!}
    \label{eq:shap_pi}
\end{equation}

\noindent\textbf{Using SHAP to explain non-intrepetable features.}  In the many cases where $x$ is not interpretable by humans, explainers such as SHAP can also provide explanations with respect to other interpretable input $x'$. In this case (which was described in detail in a recent study \cite{de2020human}), the background dataset $B$ (which is sampled from the non-interpretable input's domain) is transformed to the interpretable domain using a transformation function $h_x$ that maps $x$ to $x'$. Then, the transformed background dataset $B'$ is used for the creation of the coalition-specific background datasets $B^c_{x'}$ (note that in this case the coalition of features is sampled from the interpretable domain instead of the non-interpretable domain). Next, for each coalition-specific background dataset $B^c_{x'}$, a single instance is generated. The features of this instance represent the absence or presence of a feature within the coalition (in the interpretable domain). The prediction value of this instance $\hat{y}^{c}_{x'}$ is calculated as follows:
\begin{equation}
        \hat{y}^{c}_{x'} = \frac{1}{|B^{c}_{x'}|}  * \sum_{z\in B^{c}_{x'}}  f(h^{-1}_{x}(z))
        \label{eq:shap_interpretable_labeling}
\end{equation}
Finally, the explanation model $g$ is trained to minimize the loss function presented in Equation~\ref{eq:shap_loss_function_for_noninterpretable_domain} using the weight function presented in Equation~\ref{eq:shap_pi}.

\subsection{The Limitations of LIME and SHAP}
While LIME and SHAP have many advantages, they both suffer from the same drawback -- In order to explain a non-interpretable feature space, users must formulate an \textbf{invertible} transformation $h_x$ which maps the model's (non-interpretable) input features $x$ to the human interpretable feature space $x'$. 
As presented in Equations~\ref{eq:lime_loss_function_for_interpretable_domain} and~\ref{eq:shap_interpretable_labeling}, the original model $f$ cannot provide predictions on samples drawn from the interpretable domain $x'$ without applying the inverse transformation $h^{-1}_x$.  
However, in many real-world use cases practitioners would most likely prefer transformations that are inherently un-invertible. 
An example for a class of such transformations is transformations used for abstraction and aggregation.
These kinds of transformations are very useful for mapping a complex non-interpretable feature space to an abstract human-interpretable feature space. On the other hand, these kinds of transformations will likely map multiple low-level input features to the same high-level abstract feature. Therefore, they are inherently un-invertible. To address these limitations, we introduce Latent SHAP, which can produce human-interpretable explanations when \emph{just a non-invertible transformation function} from the model's input features to the interpretable data representation is available.
\section{\label{sec:method}Latent SHAP}
Latent SHAP extends the concept of SHAP to support the generation of practical human-interpretable explanations, in use cases where an invertible transformation function does not exist. 
The rationale behind Latent SHAP is the existence of a statistical relationship between the non-interpretable input space and the human-interpretable input space. 
Latent SHAP models this statistical relationship and eliminates the need for an invertible transformation function.

\subsection{From SHAP to Latent SHAP}
Similar to SHAP, Latent SHAP explains the predictions $f(x)$ of a complex model $f$ for a given input sample $x$ using an explanation model $g \in G$, where $G$ is a class of interpretable models. 
Additionally, both methods utilize the concept of ``Shapley values" as a measure of feature importance. 
However, the main difference between SHAP and Latent SHAP is in their ability to explain a human-interpretable input $x'$ that is different from the input used to train $f$. 
Concretely, unlike SHAP, Latent SHAP only requires a \textbf{one-way} transformation function $h_x$ that maps non-interpretable input $x$ to human-interpretable input $x'$.

\subsection{Assumptions}
Latent SHAP is able to eliminate the need for an invertible transformation function by assuming that there is a statistical relationship between the non-interpretable input $x$ and the human-interpretable input $x'$. 
Latent SHAP models this statistical relationship and estimates the predictions of $f$ for the interpretable input samples $B^c_{x'}$ (denoted as $y^c_{x'}$ in Equation~\ref{eq:shap_interpretable_labeling}) by modeling the interpretable input samples $B^c_{x'}$ as a linear combination of samples from $B_{x}$.

\subsection{Algorithm} \label{subsec:our method}
The Latent SHAP algorithm include the following eight main phases (see Figure \ref{fig:LatentShap}):

\begin{figure*}
     \centering
     \begin{subfigure}[t]{\textwidth}
         \centering
         \includegraphics[width=\textwidth]{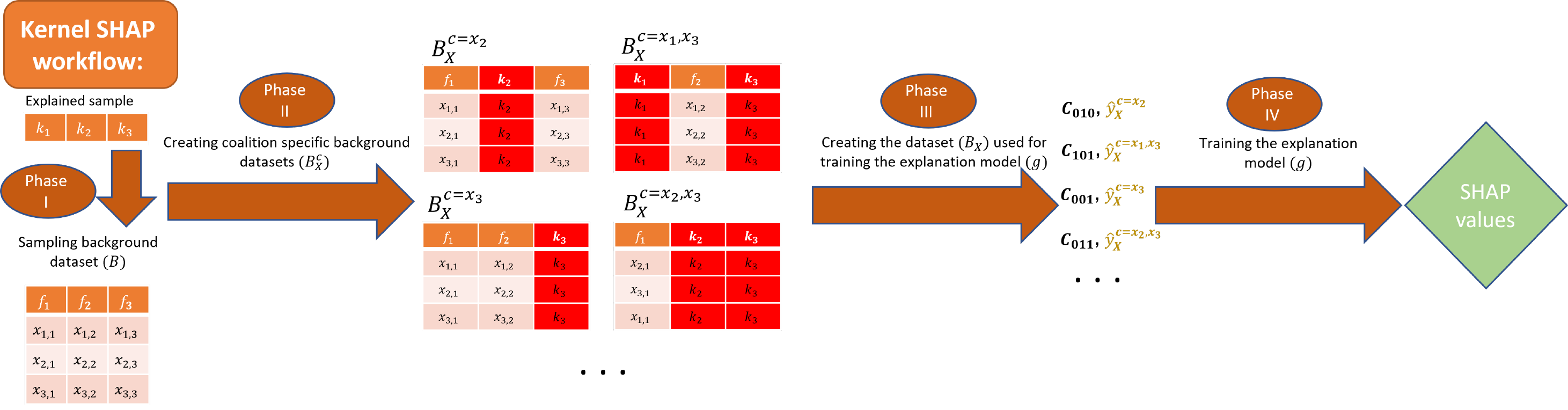}
        \caption{Kernel SHAP}
         \label{fig:KernelShap}
     \end{subfigure}
     \hfill
     \begin{subfigure}[t]{\textwidth}
         \centering
         \includegraphics[width=\textwidth]{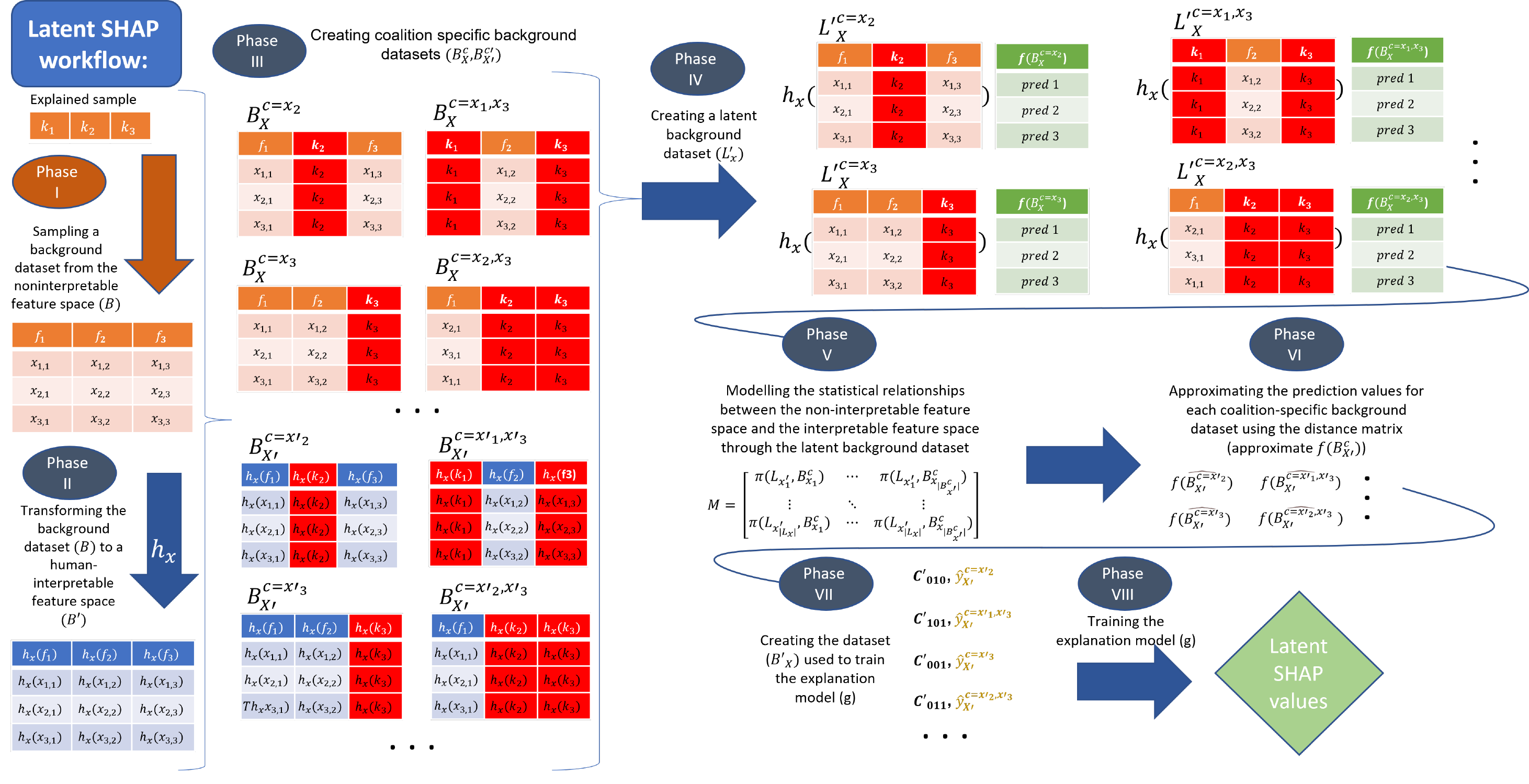}
         \caption{Latent SHAP}
         \label{fig:LatentShap}
     \end{subfigure}
        \caption{Kernel SHAP and Latent SHAP workflows.}
        \label{fig:workflows}
\end{figure*}

\noindent\textbf{\underline{Phase I:} Sampling a background dataset from the non-interpretable feature space.} Similar to SHAP, Latent SHAP's first phase is sampling a background dataset $B$ with the distribution of the dataset used to train $f$.

\noindent\textbf{\underline{Phase II:} Transforming the background dataset to a human-interpretable feature space.} The background dataset $B$ (sampled from the non-interpretable input domain) is then transformed to the interpretable domain using a transformation function $h_x$ that maps $x$ to $x'$.
We refer to this dataset as the transformed background dataset ($B'$).

\noindent\textbf{\underline{Phase III:} Creating coalition-specific background datasets.} Similar to SHAP, for each subset of features $c \subset \{x_1,x_2,...,x_n\}$ (the coalition of features), Latent SHAP creates a coalition-specific background datasets;  $B^{c}_x$, in which values of features from the background dataset $B$ that are included in the coalition $c$, are replaced with their feature values from $x$.
However, in contrast to SHAP, Latent SHAP also creates coalition-specific datasets from the transformed background dataset $B'$.
In this case, for each subset of interpretable features $c' \subset \{x'_1,x'_2,...,x'_n\}$, Latent SHAP creates a coalition-specific background dataset. 
$B^{c'}_{x'}$, in which values of features from the transformed background dataset $B'$ that are included in the coalition $c'$, are replaced with their feature values from $x'$.

\noindent\textbf{\underline{Phase IV:} Creating a latent background dataset.} In order to eliminate the need for an invertible transformation function, Latent SHAP creates a latent background dataset ($L_x'$). 
This dataset is used to model the statistical relationships between the non-interpretable inputs $x$ and the human interpretable input $x'$.
The features of this dataset are the result of applying the transformation function $h_x$ on each of the coalition-specific background datasets ($B^{c}_x$), and the prediction values of samples in $L_x$ ($\hat{Y}_{L_x}$) are calculated by applying $f$ on $B^{c}_x$ as follows:
\begin{equation}
    \begin{array}{cc}
    L_{x'} = \bigcup_{i=1}^{k} h_x(B^{c_i}_x) & \hat{Y}_{L_x} = \bigcup_{i=1}^{k} f(B^{c_i}_x)
    \end{array}
\end{equation}
where $k$ is the number of coalitions in the non-interpretable feature space, $h_x(B^{c_i}_x)$ is the application of $h_x$ to each sample in $B^{c_i}_x$, and $f(B^{c_i}_x)$ is the application of $f$ to each sample in $B^{c_i}_x$.

\noindent\textbf{\underline{Phase V:} Modeling the statistical relationships between the non-interpretable feature space and the interpretable feature space through the latent background dataset.} To model the statistical relationships between the non-interpretable feature space and the interpretable feature space, Latent SHAP calculates a distance matrix $M$ between samples in the latent background dataset $L_x$ and samples in the coalition-specific background datasets $B^{c'}_{x'}$:
\begin{equation}
        M =  
        \begin{bmatrix}
            \pi(L_{x'_1},B^{c}_{x_1}) & \cdots & \pi(L_{x'_1},B^{c}_{x_{|B^{c}_{x'}|}})\\
            \vdots &  & \vdots\\
            \pi(L_{x'_{|L_x|}},B^{c}_{x_1}) &\cdots& \pi(L_{x'_{|L_x|}},B^{c}_{x_{|B^{c}_{x'}|}})\\
        \end{bmatrix}
\end{equation}
where $\pi$, is a proximity function defined using a distance function $D$ (\eg, $L_2$ distance, cosine similarity) as in Eq.~\ref{eq:prox_function}. 
 
\noindent\textbf{\underline{Phase VI:} Approximating the prediction values for each coalition-specific background dataset using the distance matrix.} The prediction values for each coalition-specific background dataset (denoted as $f(B^{c}_{x'})$) are approximated using the distance matrix $M$, and the prediction values of samples in $L_x$ (denoted as $\hat{Y}_{L_x}$) are calculated as follows:
\begin{equation}
    f(B^{c}_{x'}) \approx \sigma(M^T) \cdot \hat{Y}_{L_x}
    \label{eq:latent_shap_approx}
\end{equation}
where $\sigma$ is the softmax function.

\noindent\textbf{\underline{Phase VII:} Creating the dataset used to train the explanation model.} The dataset used for training $g$ ($B_x'$) includes a single instance for each coalition-specific dataset generated from the transformed background dataset $B^{c'}_{x'}$. 
The features of this instance are binary indicators of the absence or presence of a feature within the coalition.
The prediction value of this instance $y^{c}_{x'}$ is calculated as follows: 
\begin{equation}
        \hat{y}^{c}_{x'} = \frac{f(B^c_{x'})}{|B^{c}_{x'}|}
\end{equation}
where $f(B^c_{x'})$ is approximated according to Equation~\ref{eq:latent_shap_approx}.
\noindent\textbf{\underline{Phase VIII:} Training the explanation model.}  Similar to SHAP, the explanation model $g$ is trained to minimize the loss function presented in Equation~\ref{eq:shap_loss_function_for_noninterpretable_domain}.

\begin{figure}
     \centering
     \begin{subfigure}[b]{0.45\textwidth}
         \centering
         \includegraphics[width=\textwidth]{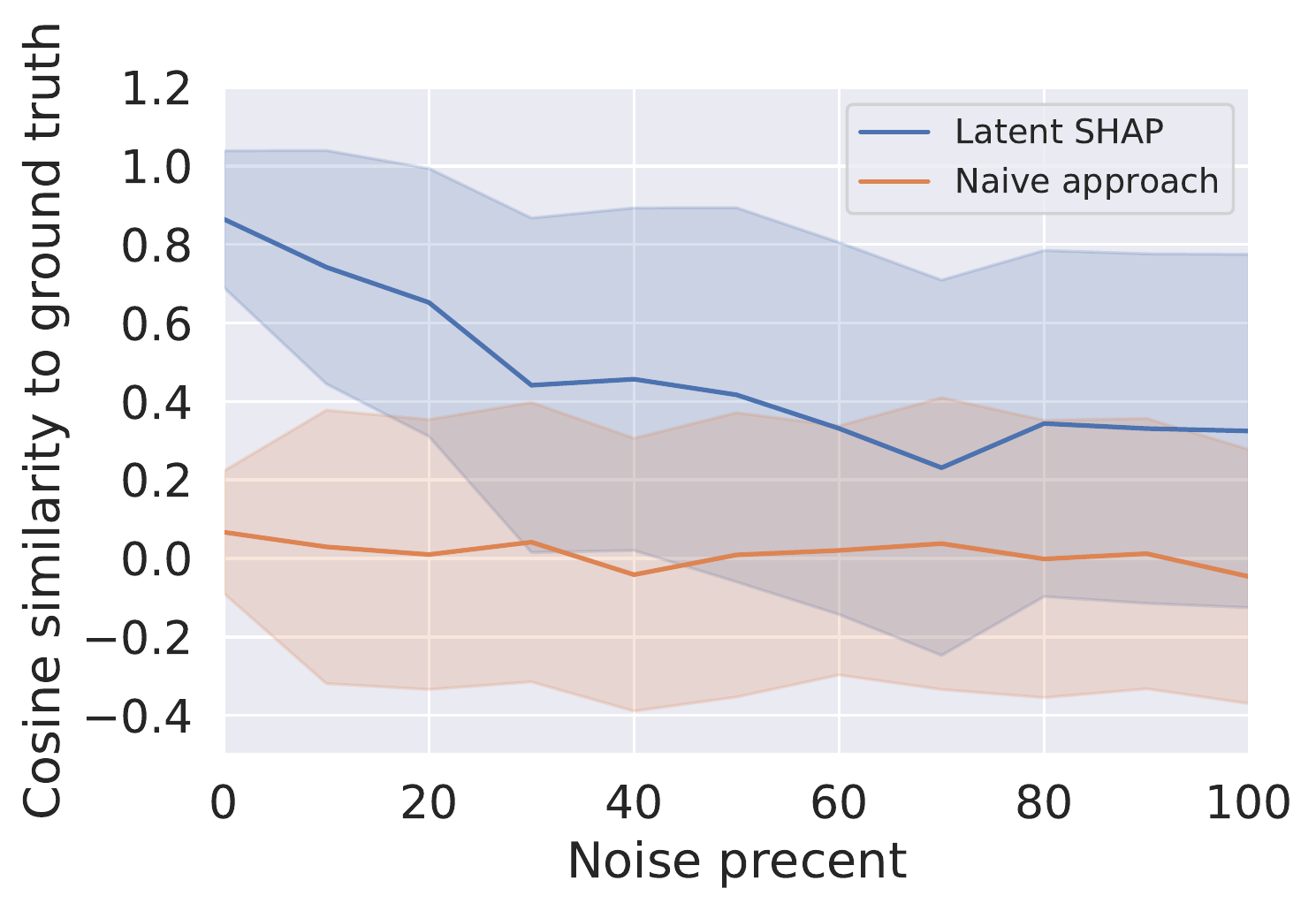}
         \caption{Background dataset size = 50}
         \label{fig:noisy_transformation_b_50}
     \end{subfigure}
     \hfill
     \begin{subfigure}[b]{0.45\textwidth}
         \centering
         \includegraphics[width=\textwidth]{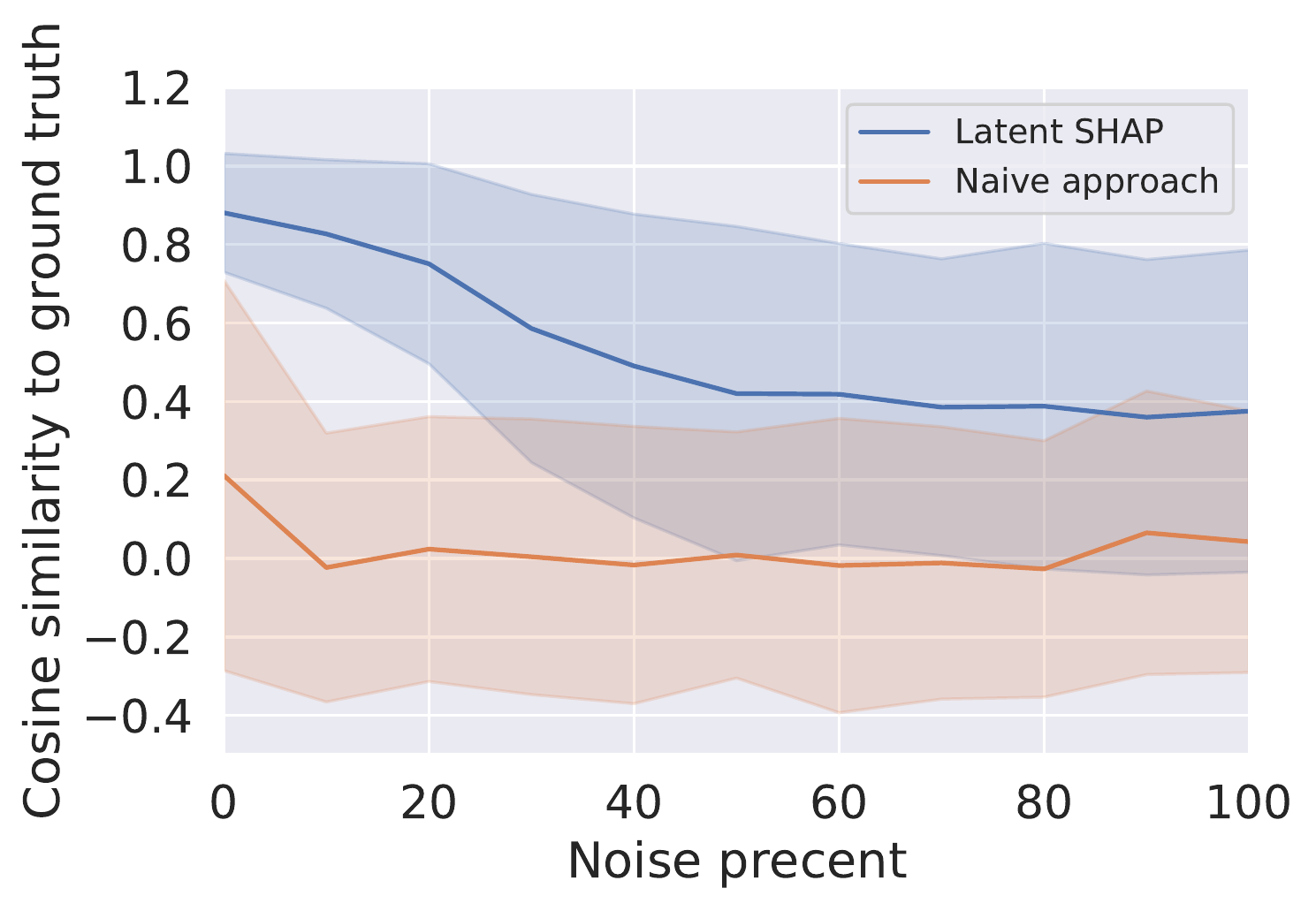}
         \caption{Background dataset size = 100}
         \label{fig:noisy_transformation_b_100}
     \end{subfigure}
     \hfill
     \begin{subfigure}[b]{0.45\textwidth}
         \centering
         \includegraphics[width=\textwidth]{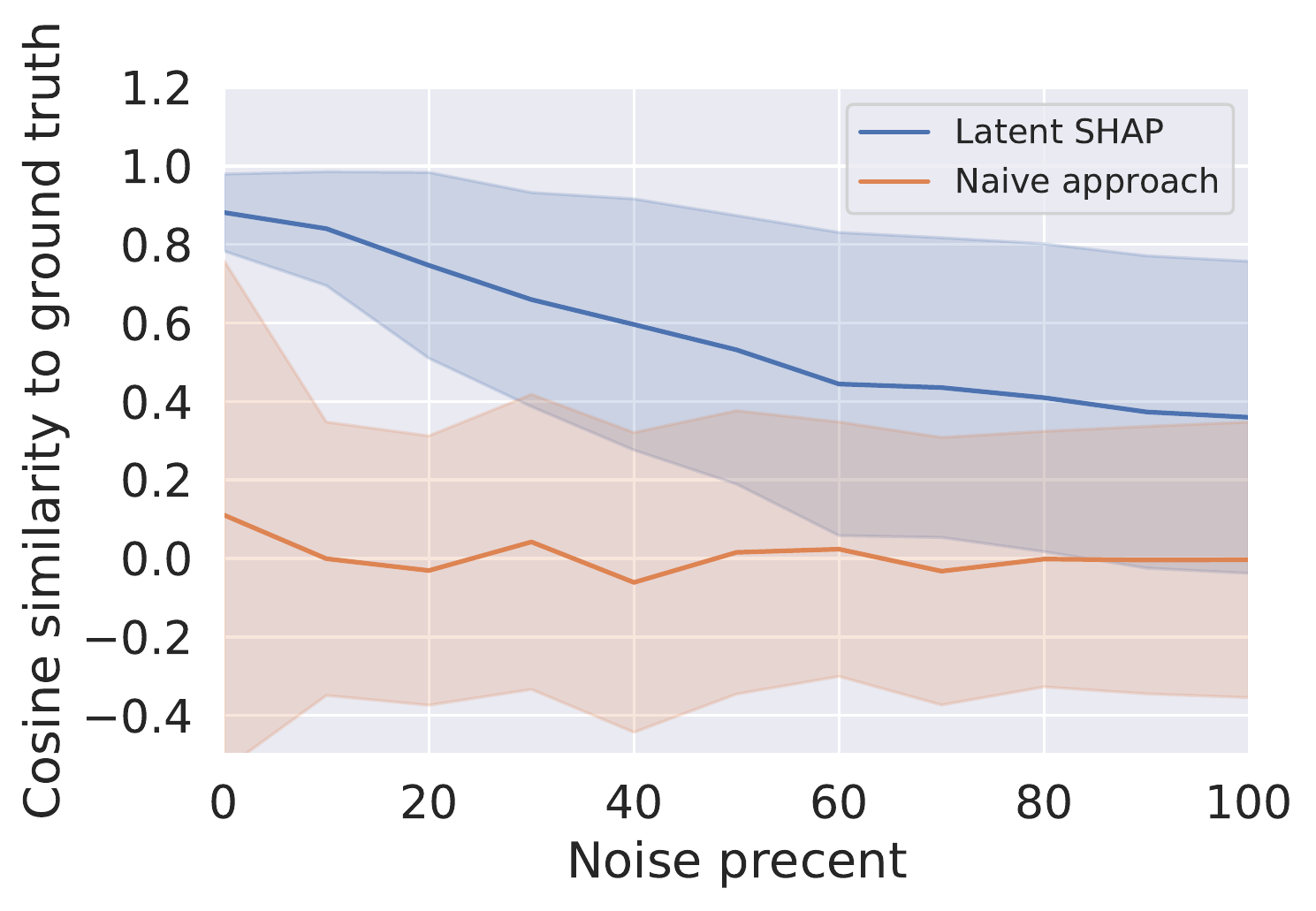}
         \caption{Background dataset size = 200}
         \label{fig:noisy_transformation_b_200}
     \end{subfigure}
        \caption{The effect of increasing noise in the transformation for different background dataset sizes.}
        \label{fig:noisy_transformation}
\end{figure}

\section{\label{sec:evaluation} Evaluation}
In this section, we evaluate Latent SHAP's performance. We conducted experiments with simulated data on a task of modeling a simple linear function (Section~\ref{sec:Controlled Evaluation})), as well as evaluation with real data on an image classification task (Section~\ref{sec:Celebrity Faces Evaluation}).

\subsection{\label{sec:Controlled Evaluation} Evaluation with Simulated Data}
In this section we address three research questions:
\noindent\textbf{\underline{RQ1:}} Are Latent SHAP's explanations locally faithful (i.e., correspond to how the model truly behaves in the vicinity of the instance being predicted).

\noindent\textbf{\underline{RQ2:}} Are Latent SHAP's explanations more locally faithful than the naive approach. \

\noindent\textbf{\underline{RQ3:}} How does a noisy transformation function affect the local faithfulness of Latent SHAP's explanations.

To answer these research questions, we performed experiments on a simple problem using synthetic data.

\subsubsection{Experimental Setup }
\noindent\textbf{\underline{The Problem:}} We begin with the simple problem of explaining a classifier $f$ that is trained to model the following function:

\begin{equation}
g(x_1,...,x_8) =
\begin{cases} 
      1 & \sum_{i=1}^{4} x_i - \sum_{i=5}^{8} x_i >0\\
      0 & otherwise
\end{cases}
\end{equation}

\noindent\textbf{\underline{The Data:}} 1,000 samples were drawn from a multivariate normal distribution parameterized by:

\begin{equation}
\centering
\begin{array}{cc}
        \Sigma =  
        \begin{bmatrix}
            \sigma_{1,1} & \cdots & \sigma_{1,8}\\
            \vdots &  & \vdots\\
            \sigma_{8,1} & \cdots & \sigma_{8,8}\\
        \end{bmatrix}
    & \mu=\vec{0}
    
\end{array}
\end{equation}
where $\sigma_{i,j}$ was sampled uniformly with $a=1,b=12$. 

\noindent\textbf{\underline{The Classifier:}} We use a simple logistic regression classifier. The classifier was trained on 500 samples and was tested on the remaining 500 samples. We observed a high test accuracy of 98.2\%.

\noindent\textbf{\underline{The Transformation Function:}} We use principal component analysis (PCA) \cite{abdi2010principal} as the transformation function. The reasons for selecting PCA, are twofold:
\begin{enumerate}
    \item PCA is an invertible transformation, therefore we can evaluate Latent SHAP's performance with respect to a ground truth (i.e, applying SHAP to explain $f$ when using the inverse of the transformation function).
    
    \item PCA is a perfect transformation (when all principal components are used), therefore we can evaluate Latent SHAP's performance with different noise ratios, by adding controlled noise to the transformation.
\end{enumerate}

\noindent\textbf{\underline{The Ground Truth:}} To obtain the ground truth, we use a Kernel SHAP explainer that utilizes the inverse PCA transformation to provide explanations with respect to the human-interpretable domain.

\noindent\textbf{\underline{The Baseline Method:}} We evaluate Latent SHAP against a naive approach, which uses Kernel SHAP for providing explanations on the non-interpretable input space and then transforms the explanations (i.e., SHAP values) to the human interpretable space using the transformation function. 

\noindent\textbf{\underline{Evaluation Metric:}} We use the cosine similarity metric to evaluate the similarity of the explanations generated by Latent SHAP (or the baseline method) to the ground truth explanations.

\subsubsection{Evaluation Results}
\noindent\textbf{\underline{RQ1:}} To answer the first research question, we measure the cosine similarity between the explanations generated by Latent SHAP and the explanations generated by the ground truth. The results are presented in Figure \ref{fig:noisy_transformation}.
As can be seen, when the background dataset is large enough (200 examples), and the transformation function is not noisy (x=0), the explanations generated by Latent SHAP are very similar to the ground truth explanations (mean cosine similarity of 0.93 over 500 testing examples). The practical insight that can be derived from this observation is that Latent SHAP can provide locally faithful explanations with respect to the non-interpretable feature space when the transformation function is accurate. 

\noindent\textbf{\underline{RQ2:}} To answer the second research question, we also measure the cosine similarity between the explanations generated by the baseline method and the explanations generated by the ground truth. The results are presented in Figure \ref{fig:noisy_transformation}. As can be seen, the naive approach yields very poor performance. The practical insight that can be derived from this observation is that the naive approach provides explanations that are not locally faithful.

\noindent\textbf{\underline{RQ3:}} To answer the third research question, we evaluate the explanations generated by the baseline and Latent SHAP methods when Gaussian noise is added to the output of the transformation function.
Specifically, for each feature in the interpretable input space (denoted as $x'_i$) we added Gaussian noise with $\mu=0$ and $\sigma = \alpha * \sigma_{x'_i}$, where $\sigma_{x'_i}$ is the standard deviation of the feature $x'_i$, 
and $\alpha \in [0,1]$ is a scaling factor.
In our experiments we examined various values of $\alpha$, and the results of this examination are presented in Figure \ref{fig:noisy_transformation}. As expected, adding noise reduces our method's performance, with a slower reduction in performance seen with larger background dataset sizes. Nevertheless, Latent SHAP demonstrated high performance (mean cosine similarity of 0.9-0.78) for small noise ratios ($\alpha < 0.15$) when the background dataset is large enough (size $>$ 100). Based on these findings we conclude that Latent SHAP can provide locally faithful explanations with respect to the non-interpretable feature space when the transformation function is relatively accurate. 
\begin{figure}[t]
    \centering
    \includegraphics[width=0.5\textwidth]{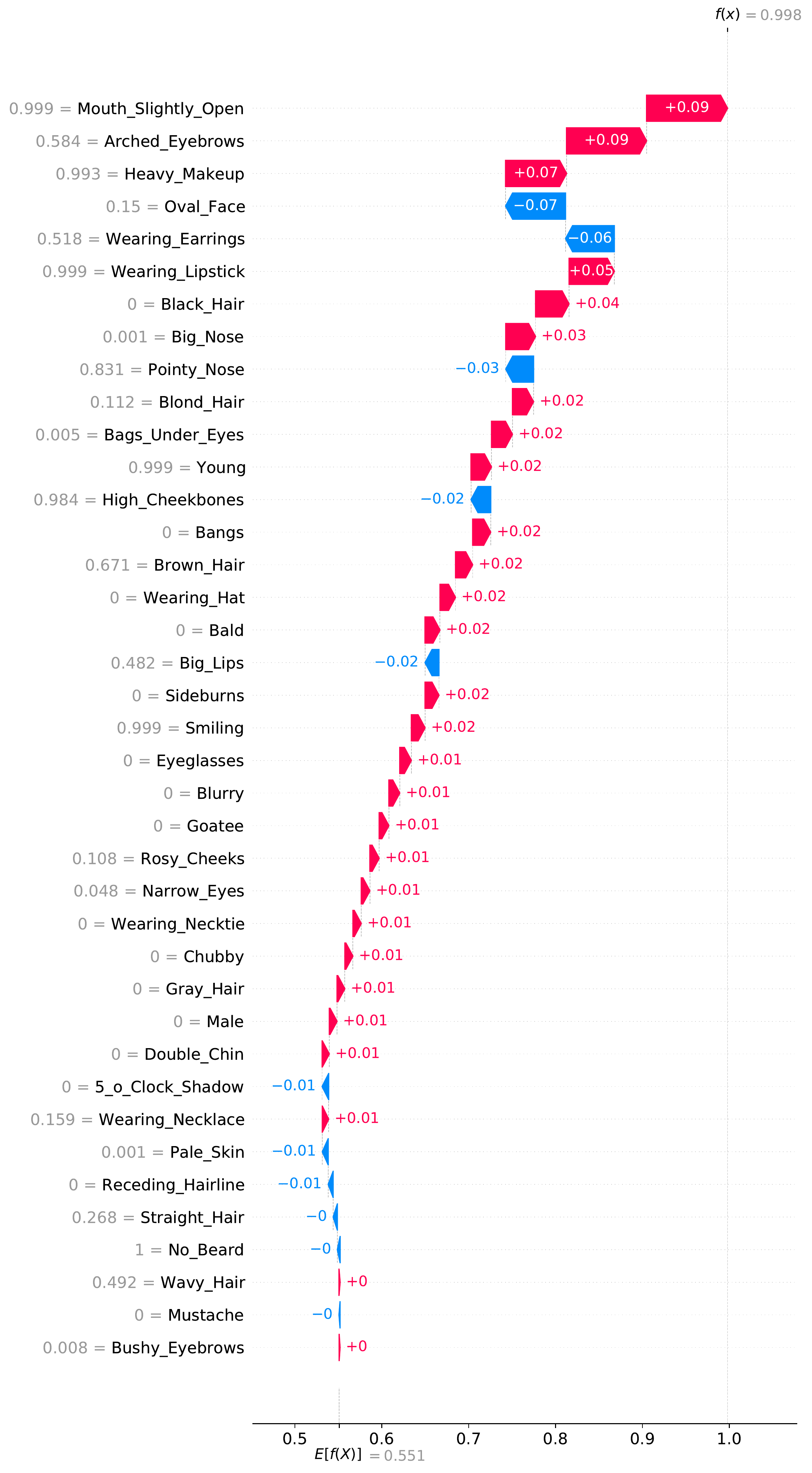}
    \caption{Attractiveness Local explanation for all human-interpretable features for the image in Figure 1.}
    \label{fig:all features local explanation}
\end{figure}
\subsection{\label{sec:Celebrity Faces Evaluation}Evaluation with Real Data}
In this we address the following research questions:
\noindent\textbf{\underline{RQ 1:}} Do Latent SHAP's local explanations enable better understanding of model predictions than those provided by SHAP?

\noindent\textbf{\underline{RQ 2:}} Are the explanations generated by Latent SHAP logical on a global level?
\subsubsection{Experimental Setup}
\noindent\textbf{\underline{The Problem:}} We continue with a real-world problem of explaining a classifier $f$ used for image classification. Specifically, the classifier is trained to predict the attractiveness of celebrities' based on their image faces.

\noindent\textbf{\underline{The Data:}} We use the CelebA dataset \cite{liu2018large} which contains 200,00 images (200k) of celebrities' faces annotated with 40 binary human-interpretable features, that were manually labeled (such as wearing hat, smiling, pointy nose).
One of these feature is the attractiveness of the face of the celebrity in the image. In our evaluation, we consider the attractiveness feature as the label and the rest of the features as human-interpretable attributes.

\noindent\textbf{\underline{The Classifier:}} We train a convolutional neural network (CNN) classifier to predict the attractiveness of images of celebrities.
The classifier's architecture include 13 convolutional layers and a single Linear layer. Each convolutional layer was followed by a batch normalization layer, and a ReLU activation function. Max-pooling layers follow the second, fourth, seven, and tenth, and thirteen  convolutional layers. Dropout layers  follow the tenth and  thirteen convolutional layers.
The model was optimized using the momentum optimizer (learning rate= 0.1, momentum= 0.9) and the binary-cross-entropy loss. The classifier was trained on 180k samples and was validated and tested on the remaining 20k samples. We observed a test f1 score of 78.5\%.

\noindent\textbf{\underline{The Transformation Function:}}
We train another deep neural network classifier to predict the other 39 human-interpretable features given images of celebrities. The classifier's architecture was similar to $f$
The model was optimized using the momentum optimizer (learning rate= 0.1, momentum= 0.9) and a binary-cross-entropy loss. This classifier was trained and tested as the above classifier. We observed a test f1 score of 82.7\%.

\noindent\textbf{\underline{Baseline Methods:}}
We use the SHAP Partition explainer, which commonly used for explaining high-dimensional data, as the baseline method.
SHAP partition explainer approximates Shapley values in polynomial time (instead of exponential), by using Owen values. The explanations generated by SHAP Partition explainer colors groups of pixels that increase the attractiveness prediction in red and pixels that reduce the attractiveness prediction in blue, with the color saturation indicating how much a pixel contributes.

\subsubsection{Evaluation Results}
\noindent\textbf{RQ1:} To answer the first research question, we evaluate Latent SHAPs explanations qualitatively. In Figure \ref{fig:all features local explanation}, we present Latent SHAP's explanation for the image face presented in Figure \ref{fig:intro}. As can be noticed, Latent SHAP's explanations enable high-level understanding of the prediction model, which is one of the main goals of explanation frameworks. The explanations presented in Figure \ref{fig:all features local explanation} indicate that most of the features have little or no impact, which indicate the lack of a facial feature.

\begin{figure*}[t]
    \centering
    \begin{tabular}{cccc}
    Explained image & SHAP & Latent SHAP (Background = 50) & Latent SHAP (Background = 150) \\
    \includegraphics[width=0.14\textwidth, height=0.2\textwidth]{figures/celebA/explained_image.png} &
    \includegraphics[width=0.14\textwidth, height=0.2\textwidth]{figures/celebA/local_eplanation_1_pixel_level.png}&
    \includegraphics[width=0.36\textwidth, height=0.2\textwidth]{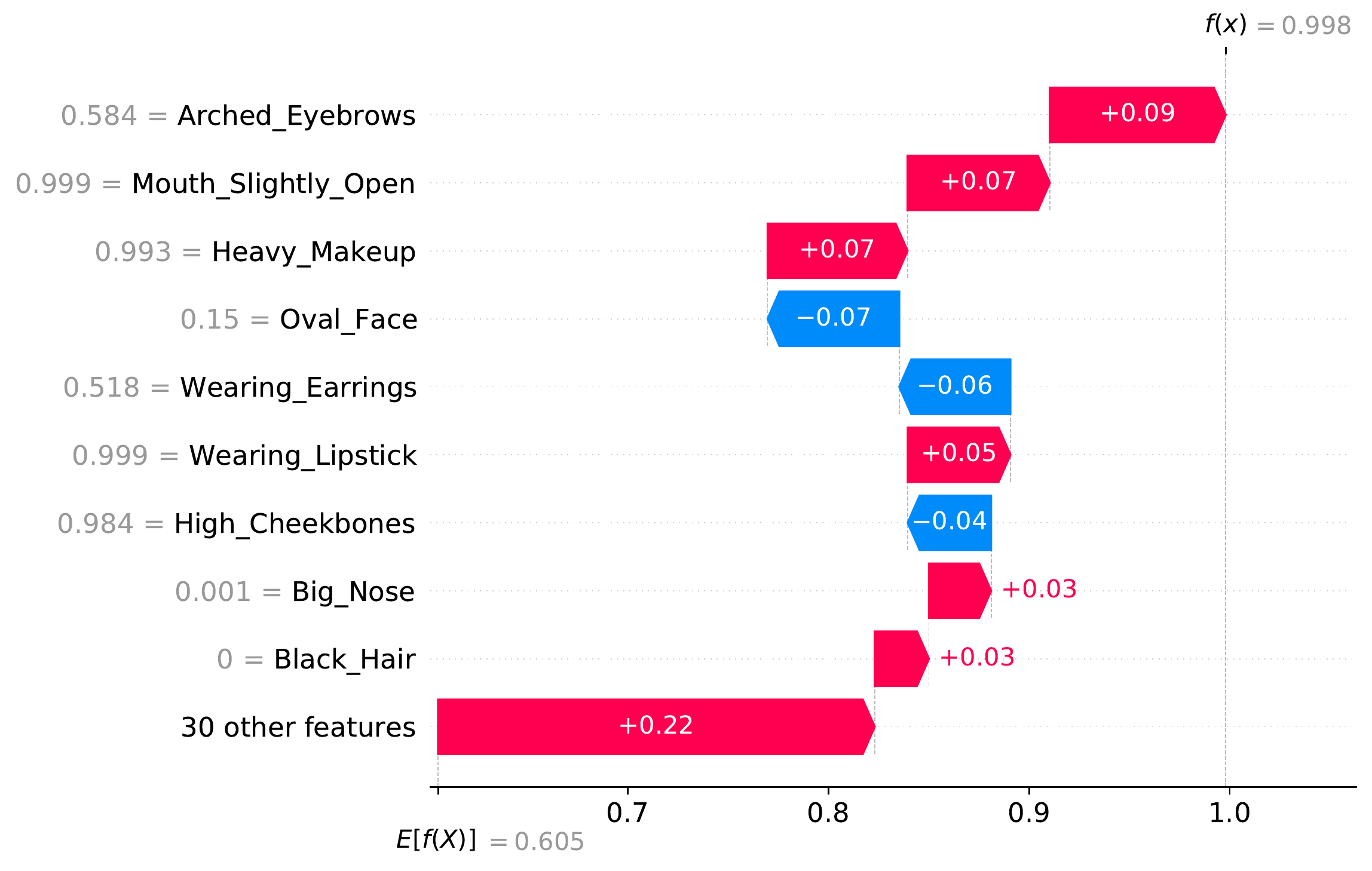}&
    \includegraphics[width=0.36\textwidth, height=0.2\textwidth]{figures/celebA/bg_size_150/local_eplanation_1_bg_150.pdf}
    \\ 
    \includegraphics[width=0.14\textwidth, height=0.2\textwidth]{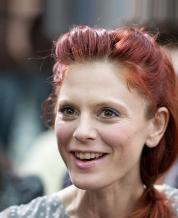}     &
    \includegraphics[width=0.14\textwidth, height=0.2\textwidth]{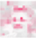}    &
    \includegraphics[width=0.36\textwidth, height=0.2\textwidth]{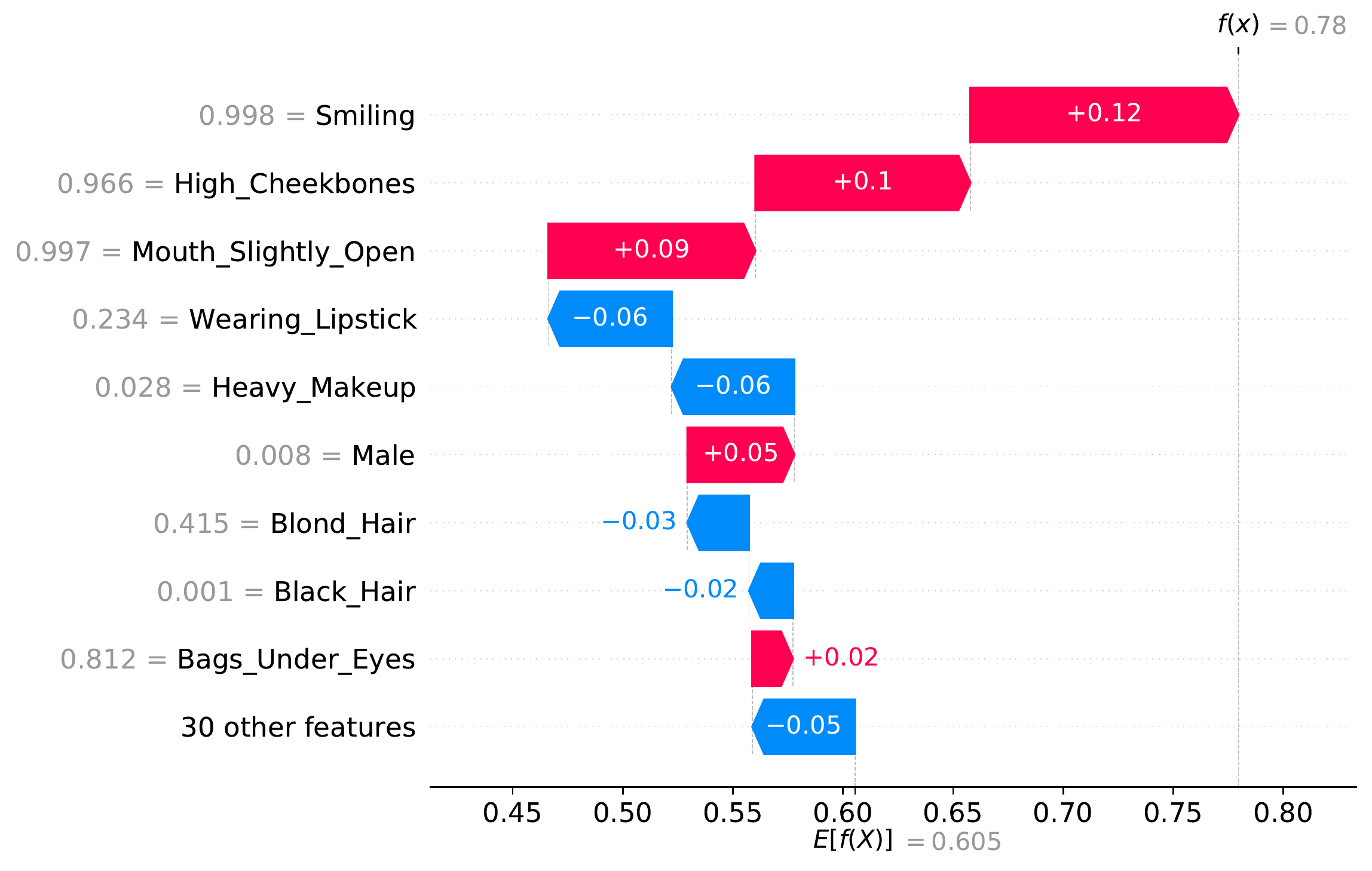}   &
    \includegraphics[width=0.33\textwidth, height=0.2\textwidth]{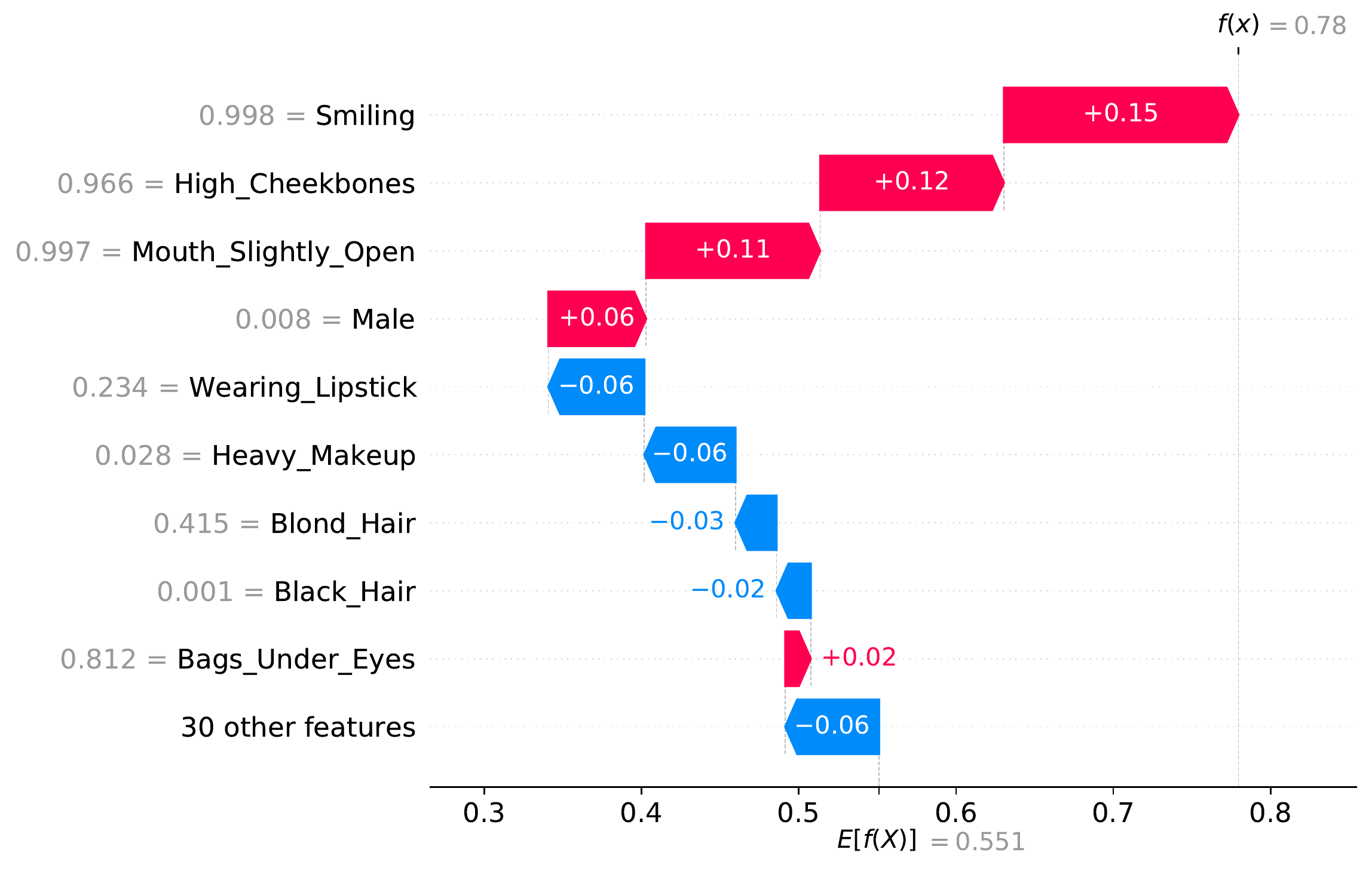} 
    \\ 
    \includegraphics[width=0.14\textwidth, height=0.2\textwidth]{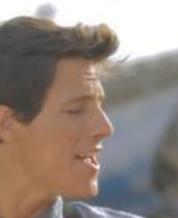} &
    \includegraphics[width=0.14\textwidth, height=0.2\textwidth]{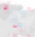}&
    \includegraphics[width=0.36\textwidth, height=0.2\textwidth]{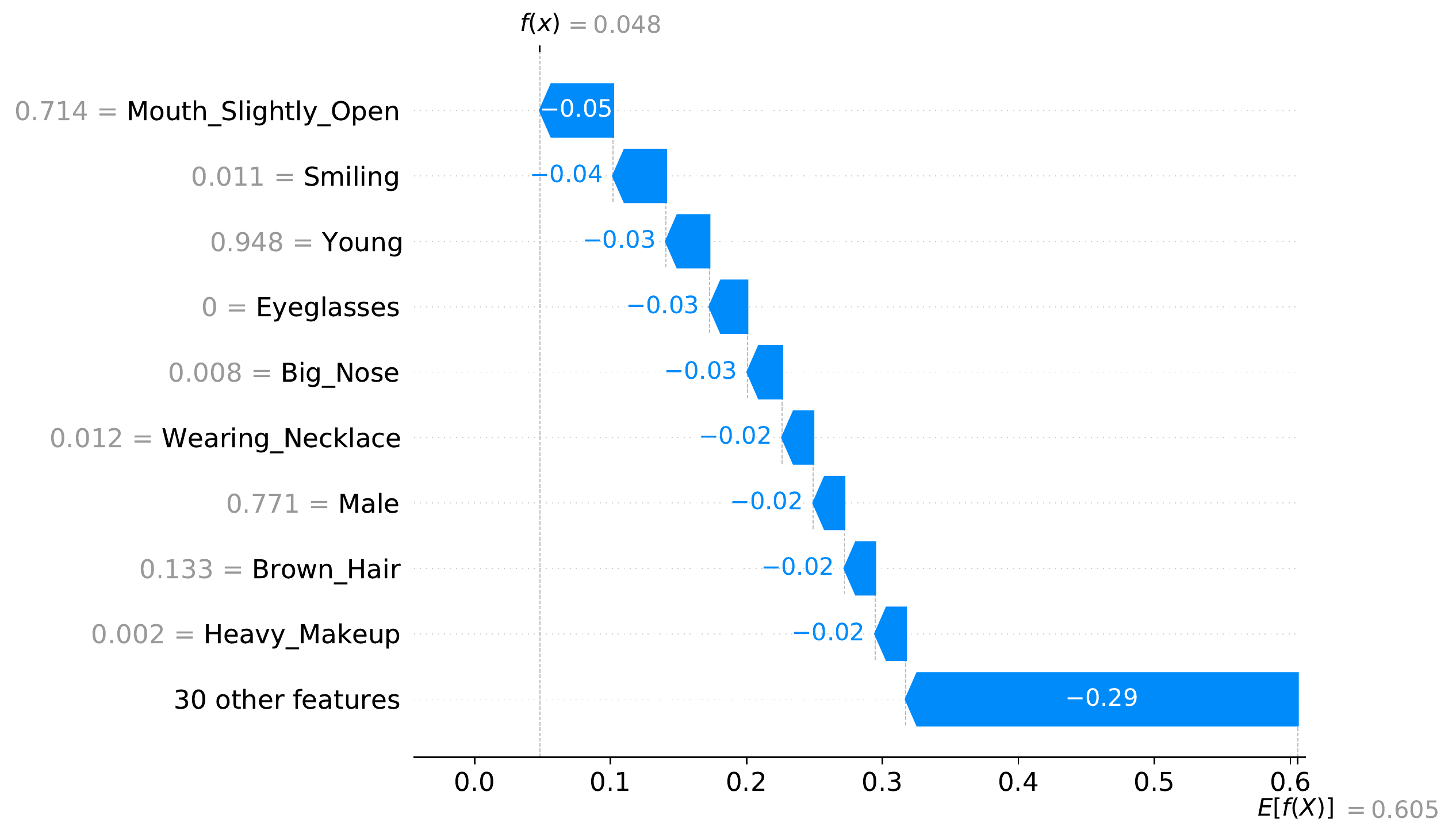}&
    \includegraphics[width=0.36\textwidth, height=0.2\textwidth]{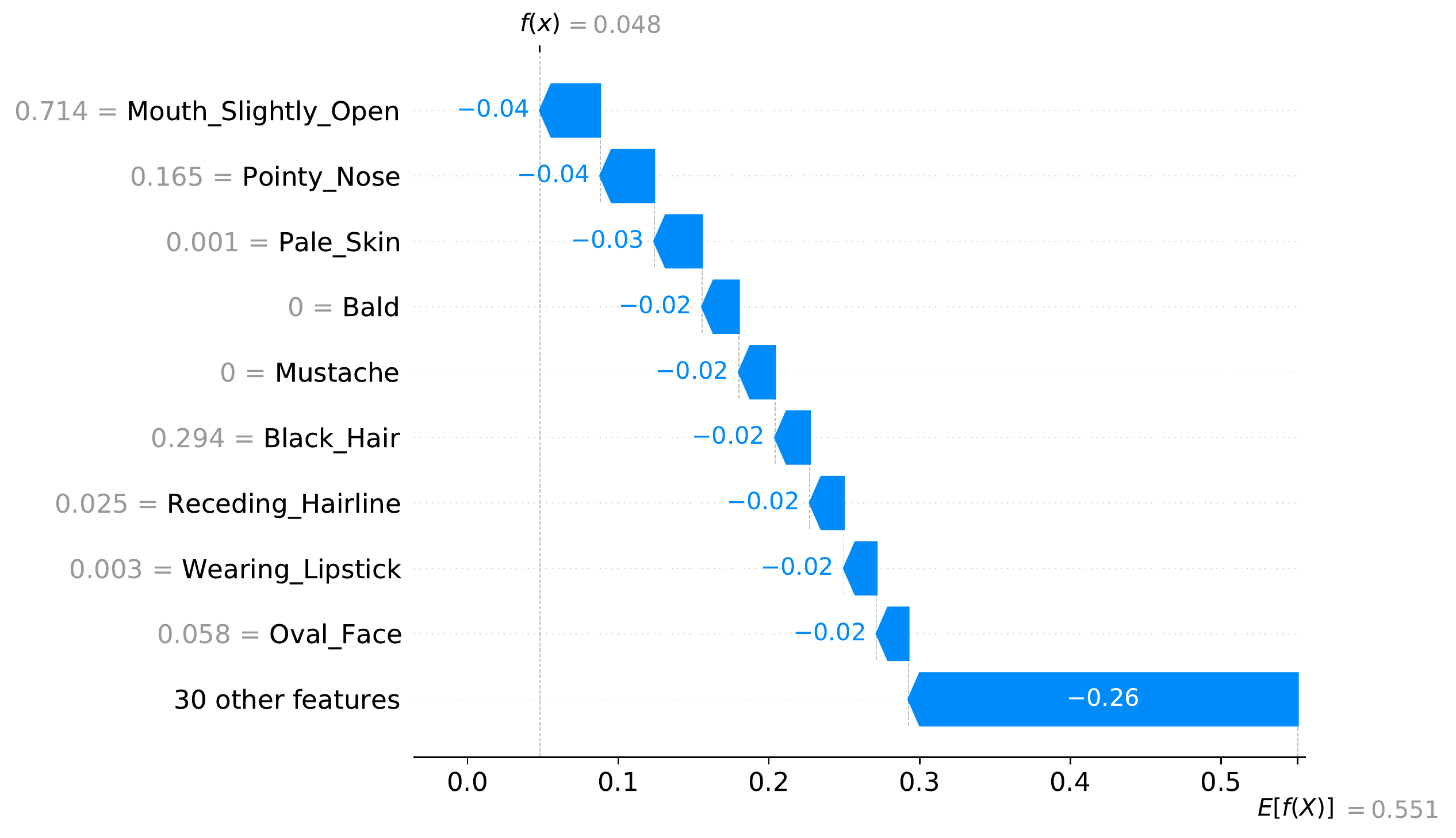}
    \\ 
    \includegraphics[width=0.14\textwidth, height=0.2\textwidth]{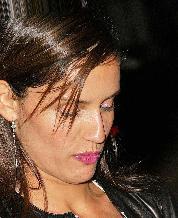}&
    \includegraphics[width=0.14\textwidth, height=0.2\textwidth]{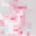}&
    \includegraphics[width=0.36\textwidth, height=0.2\textwidth]{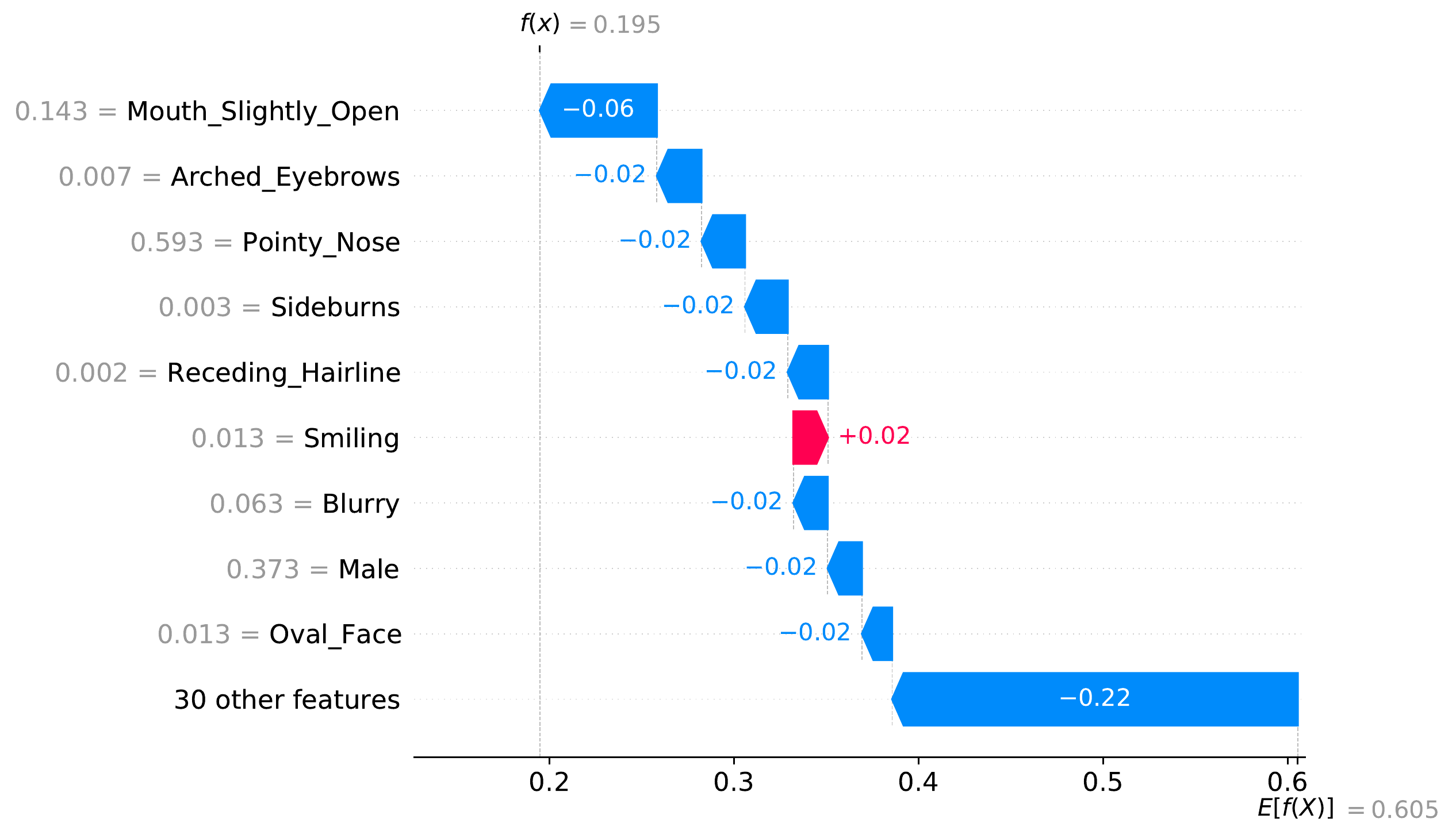}&
    \includegraphics[width=0.36\textwidth, height=0.2\textwidth]{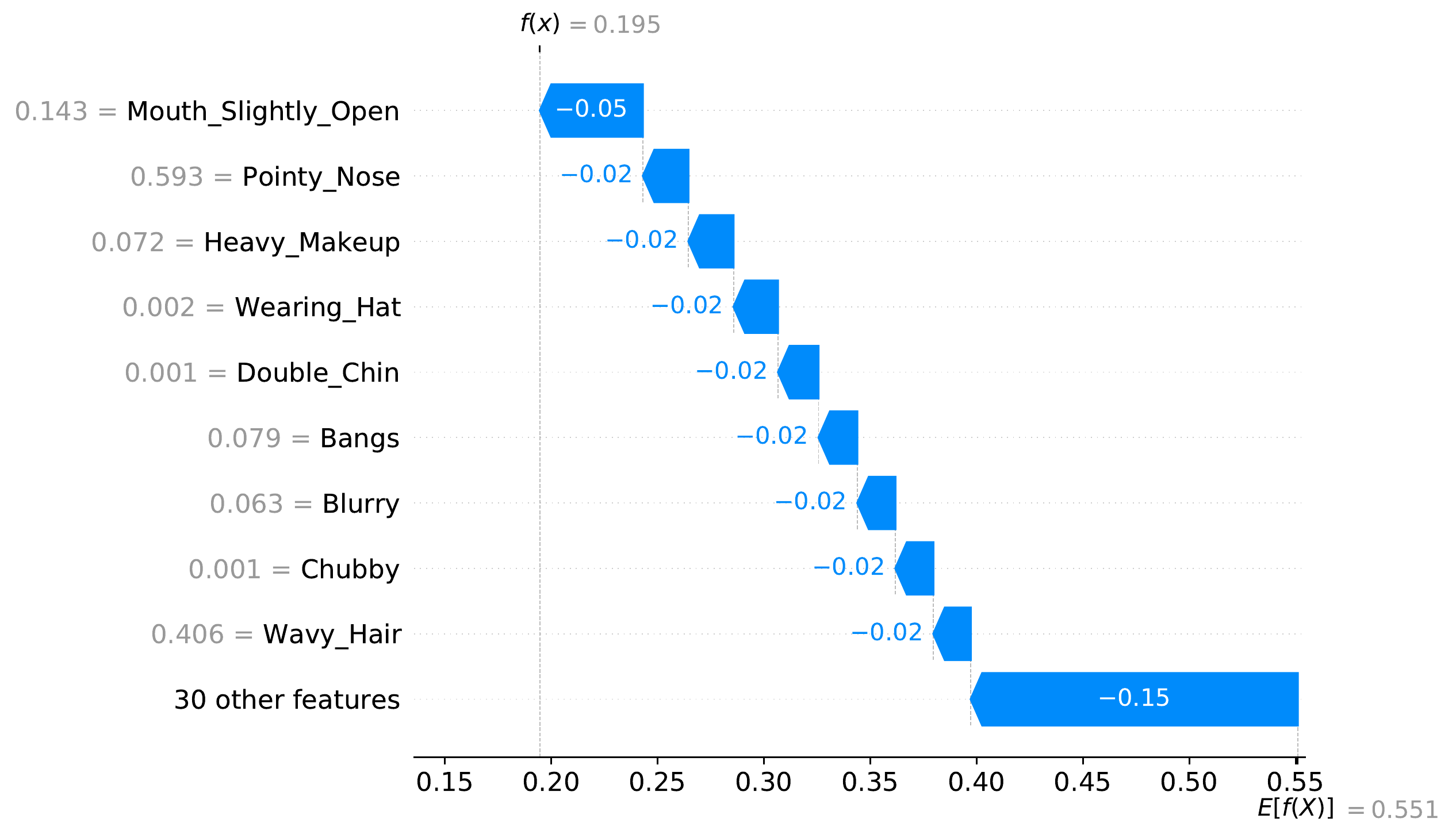}
    \end{tabular}
    \caption{Local explanations for celebrity attractiveness prediction: SHAP vs. Latent SHAP with different background dataset sizes}
    \label{fig:local explanation examples}
\end{figure*}

In Figure \ref{fig:local explanation examples}, we present four local Latent SHAP explanations for two facial images predicted to be attractive and two predicted to be unattractive.
Latent SHAP provides explanations that are more informative than SHAP's pixel-based explanations and makes it easier to draw conclusions about the prediction model.
For example, some concepts may be difficult to identify when highlighted in the pixel space (such as smiling, open mouth, face shape, hair color), but when the image is transformed to the human-interpretable feature space, the concepts can be easily identified.
In general, we found that when a face is classified as unattractive, Latent SHAP assigns low negative feature importance to many features.
On the other hand, when a face is classified as attractive, Latent SHAP assigns high positive feature importance to just a few features. 

As can be seen in Figure \ref{fig:local explanation examples},  the size of the background dataset affects the results and should be selected carefully.
For example, for the first image, the contribution of the 'high cheekbones' feature diminishes, and it moves from being the seventh most contributing feature (in absolute terms) with an importance score of -0.04 to a position outside the top-9 (as its importance value gets closer to zero) when the background size is increased from 50 to 150. This decrease in importance is significant enough that when 150 background samples are used, the 'high cheekbones' feature is no longer one of the top-9 contributing features and therefore does not appear in the figure.

\setlength{\tabcolsep}{0pt}

\begin{figure*}[ht]
    \centering
    \includegraphics[width=0.9\textwidth, height=0.27\textwidth]{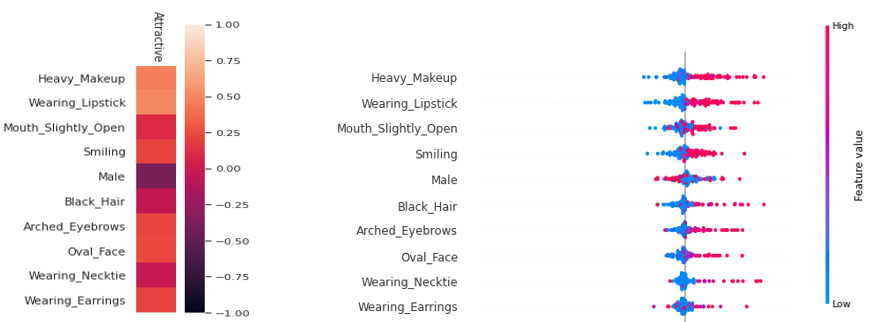}
    \caption{Feature correlation for 'attractive' vs. Latent SHAP's global explanation for the classification of celebrities' attractiveness for features with the top-10 mean absolute importance values.}
    \label{fig:corr and global explanation}
\end{figure*}

\noindent\textbf{\underline{RQ2:}} To answer the second research question, we compare the global explanations generated by Latent SHAP (Figure~\ref{fig:corr and global explanation} right) to the ground-truth correlations of the human-interpretable features for attractiveness (Figure~\ref{fig:corr and global explanation} left).
Note that we only present the 10 most influential features.
The individual Latent SHAP feature attributions, denoted as Latent SHAP values, appear as dots on the right side of Figure~\ref{fig:corr and global explanation}. 
Each dot represents a single sample, where the x-axis is the Latent SHAP value, the color represents the feature value, and the y-axis shows bunching where many samples have the same feature values.

As can be seen in Figure~\ref{fig:corr and global explanation}, globally, the Latent SHAP values correspond well to ground-truth correlations with attractiveness.
For example, 'Heavy Makeup' is positively correlated with attractiveness, and the Latent SHAP values indicate a low contribution for low feature values and a high contibution for high feature values.
Latent SHAP also assigns appropriate Latent SHAP values to the male feature which is inversely correlated with attractiveness.

\section{\label{sec:conclusion}Conclusion}
In this paper, we presented Latent SHAP, a framework for generating locally faithful model-agnostic human-interpretable explanations, given  an un-invertible transformation function from the features the model was trained on to human-interpretable features that are used in the explanation. Our experiments show that Latent SHAP's explanations are both locally faithful and highly intuitive.

{\small
\bibliographystyle{ieee_fullname}
\bibliography{LatentShap}
}
\end{document}